\pdfoutput=1

\documentclass[11pt]{article}

\usepackage[]{ACL2023}

\usepackage{times}
\usepackage{latexsym}

\usepackage[T1]{fontenc}

\usepackage[utf8]{inputenc}

\usepackage{microtype}

\usepackage{inconsolata}

\usepackage{algorithm}
\usepackage{algpseudocode}
\usepackage{xspace}
\usepackage{multicol}
\usepackage{caption}
\usepackage{subcaption}
\usepackage{graphicx}
\usepackage{amsmath}
\usepackage{hyperref}
\usepackage{amsfonts}
\usepackage{pifont}
\usepackage{makecell}
\usepackage{tablefootnote}

\newcommand{\ours}{\textsc{SenteCon}\xspace}
\newcommand{\oursp}{\textsc{SenteCon+}\xspace}
\newcommand{\oursboth}{\textsc{SenteCon(+)}\xspace}

\title{\ours: Leveraging Lexicons to Learn Human-Interpretable Language Representations}

\author{Victoria Lin \\
  Carnegie Mellon University \\
  \texttt{vlin2@andrew.cmu.edu} \\\And
  Louis-Philippe Morency \\
  Carnegie Mellon University \\
  \texttt{morency@cs.cmu.edu} \\}

\begin{document}
\maketitle
\begin{abstract}

Although deep language representations have become the dominant form of language featurization in recent years, in many settings it is important to understand a model's decision-making process. This necessitates not only an interpretable model but also interpretable features. In particular, language must be featurized in a way that is interpretable while still characterizing the original text well. We present \ours, a method for introducing human interpretability in deep language representations. Given a passage of text, \ours encodes the text as a layer of interpretable categories in which each dimension corresponds to the relevance of a specific category. Our empirical evaluations indicate that encoding language with \ours provides high-level interpretability at little to no cost to predictive performance on downstream tasks. Moreover, we find that \ours outperforms existing interpretable language representations with respect to both its downstream performance and its agreement with human characterizations of the text.
\end{abstract}

\section{Introduction}
\label{sec:introduction}

\begin{figure*}[!ht]
    \centering
    \includegraphics[width=\textwidth]{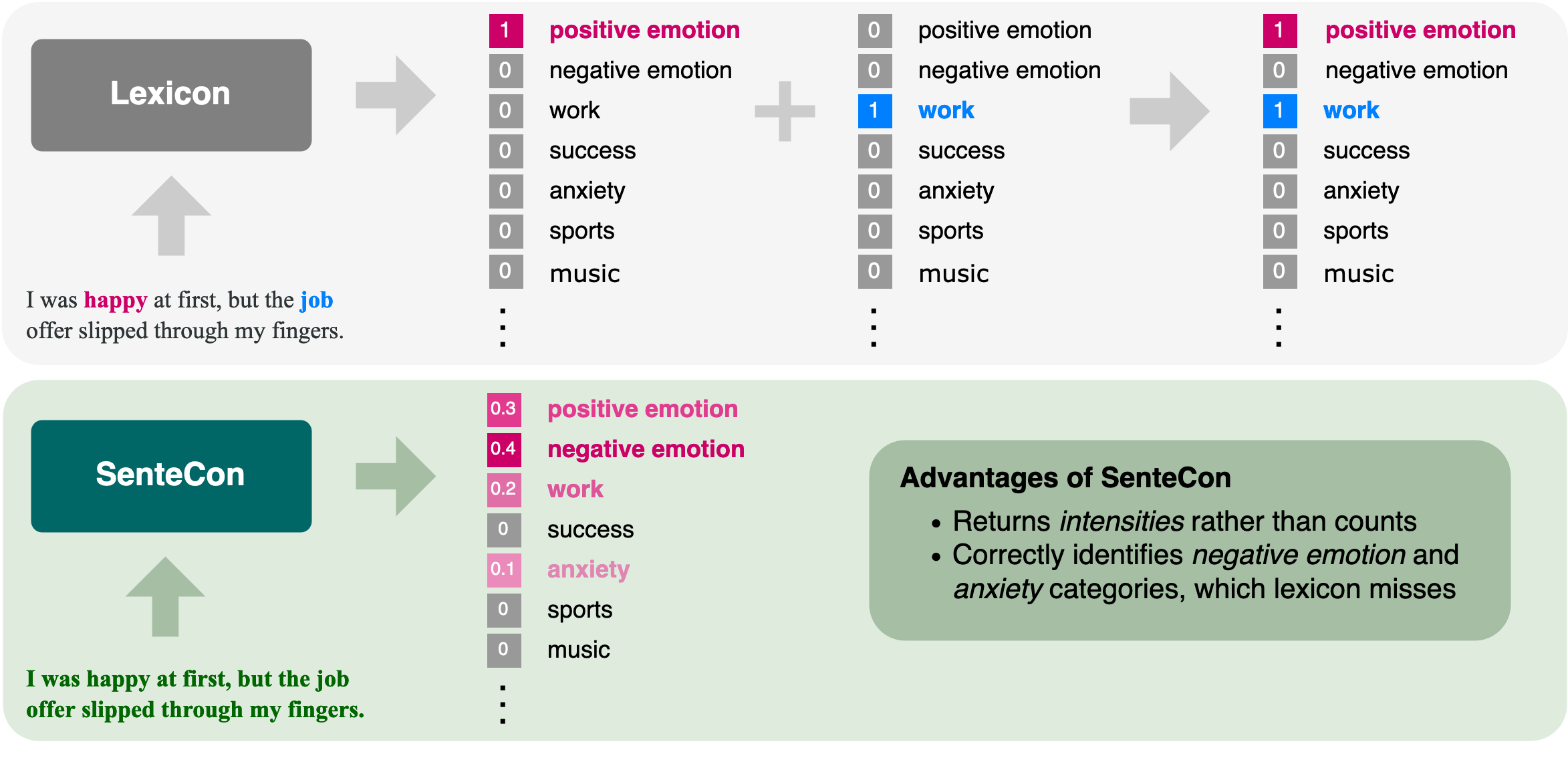}
    \caption{A comparison of lexicon-based language representations and \ours. While lexicons encode word-level category counts, \ours parses whole sentences and encodes sentence-level category intensities.}
    \label{fig:lexicon_vs_sentecon}
\end{figure*}


Deep language representations have become the dominant form of language featurization in recent years. These black-box representations perform excellently on a diverse array of tasks and are widely used in state-of-the-art machine learning pipelines. In many settings, however, it is important to understand a model's decision-making process, which necessitates not only an interpretable model but also interpretable features. To be useful, language must be featurized in a way that is interpretable while still characterizing the original text well. The fields of affective computing, computational social science, and computational psychology often use models to elucidate the relationships between patterns of language use and specific outcomes \citep{lin2020toward, wortwein2021human}. Moreover, interpretability is necessary to enforce desirable criteria like fairness \citep{du2021fairness}, robustness \citep{doshi2017towards}, and causality \citep{veitch2020adapting, feder2022causal}.


Despite advances in deep language representations, they are not considered human-interpretable due to their high dimensionality and the fact that their dimensions do not correspond to human-understandable concepts. Instead, researchers in need of interpretable language representations often turn to lexicons \citep{morales2017cross, saha2019social, relia2019race}, which map words to meaningful categories or concepts. While useful in their simplicity, lexicons capture much less information about the text than do deep language representations. Most notably, because they parse text on the level of individual words, lexicons are unable to represent how those words are used within the broader context of the text, which can lead to misrepresentation of the text's meaning or intent. Consequently, lexicon-based language representations may not necessarily correspond well with how a human, who is able to comprehend the entire passage context, would perceive the text; and they may not perform well when used in downstream tasks.


With an eye toward addressing these concerns, we present \ours,\footnote{Our code and data are publicly available at \url{https://github.com/torylin/sentecon/}.} a method for introducing human interpretability in deep language representations. Given a sentence,\footnote{We use the term ``sentence'' for clarity, but our approach is also applicable to longer passages of text like paragraphs and documents, as our experiments show.} \ours encodes the text as a layer of interpretable categories in which each dimension corresponds to the relevance of a specific category (Figure \ref{fig:lexicon_vs_sentecon}). The output of \ours can itself therefore be viewed as an interpretable language representation. As language use can vary across text domains, we also present an extension, \oursp, that can adapt to specific domains via a \textit{reference corpus}, a collection of unlabeled text passages from a target domain.

We evaluate \ours and \oursp (jointly denoted hereafter as \oursboth) with respect to both human interpretability and empirical performance. We first conduct an extensive human study that measures how well \oursboth characterizes text compared to traditional lexicons. We complement this study with experiments using \oursboth interpretable representations in downstream tasks, which allow us to compare its performance with that of existing interpretable and non-interpretable language representations. 
Finally, we analyze \oursboth representations to determine whether they indeed are influenced by sentence context in a meaningful way.


\begin{figure*}[!ht]
    \centering
    \includegraphics[width=0.8\textwidth]{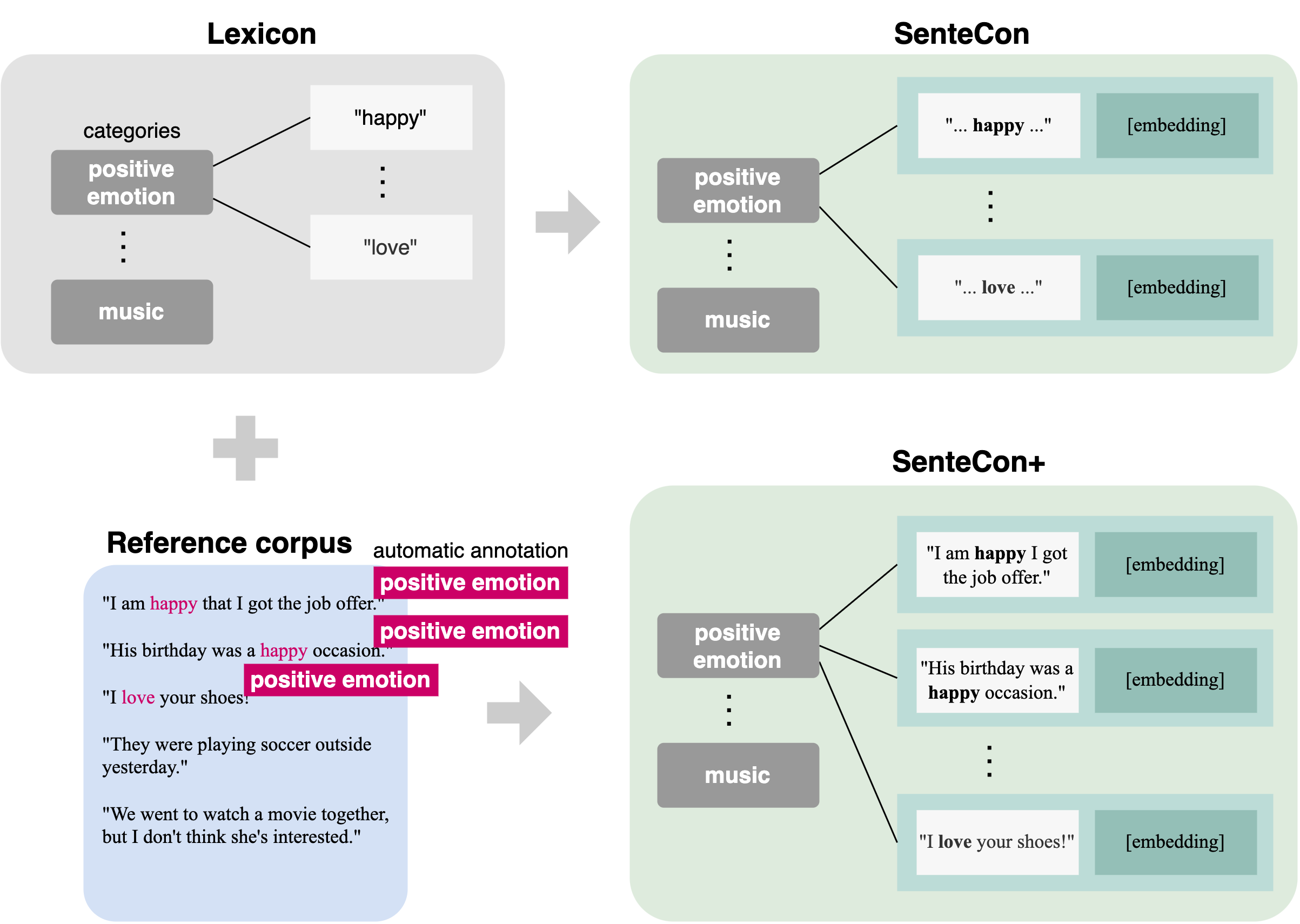}
    \caption{An illustration of the \ours and \oursp methods. Starting with a traditional lexicon, it is possible to obtain either \ours (top row) or---using a reference corpus---\oursp (bottom row).}
    \label{fig:lexicon_to_sentecon}
\end{figure*}

\section{Related Work}
\label{sec:related_work}

\textbf{Lexicons.} One of the primary existing interpretable language representations is the \textit{lexicon}. A lexicon is a mapping of words to one or more categories (often linguistic or topical) that can be used to compute a score or weight for those categories from a passage of text. Popular lexicons include Linguistic Inquiry and Word Count (LIWC), a human-constructed lexicon for psychology and social interaction \citep{pennebaker2015development}; Empath, a general-purpose lexicon in which categories are generated automatically from a small set of seed words \citep{fast2016empath}; and SentiWordNet, an automatically-generated lexicon for sentiment analysis and opinion mining \citep{baccianella2010sentiwordnet}. 



\textbf{Contextual lexicons.} Contextual lexicons attempt to incorporate sentence context while retaining a lexicon structure. In one class of methods, adjustments are made to the lexicon via human-defined rules that depend on the context of the word being parsed. However, the reliance of these rule-based approaches on human intervention limits their wider use. For example, \citet{muhammad2016contextual} modify the sentiment score output of their lexicon based on the proximity of negation words and valence shifters, and \citet{vargas2021contextual} construct a lexicon that explicitly defines words that are context-independent (i.e., will retain their meaning regardless of context) and context-dependent.


\textbf{Interpretable deep language models.} A number of works provide some degree of interpretability to black-box language models via post-hoc analyses. \citet{clark2019bert} analyze BERT's \cite{devlin-etal-2019-bert} attention heads and link them to attributes like delimiter tokens or positional offsets, while \citet{bolukbasi2021interpretability} examine individual neurons within the BERT architecture that spuriously appear to encode a single interpretable concept. \citet{gorski2020towards} adapt the Grad-CAM visual explanation method \cite{selvaraju2017grad} for a CNN-based text processing pipeline. Although these analyses lend some insight, interpretability is limited to the low-level concepts associated with individual attention heads or neurons, and substantial manual probing is required for each network.

\textbf{Methods.} Several previous works contain elements of methodological similarity to \oursboth but differ in their aims. To address gaps in the LIWC lexicon vocabulary, \citet{gong2018improving} implement a soft matching scheme based on non-contextual WordNet \citep{miller1995wordnet} and word2vec \citep{mikolov2013efficient} embeddings. Given a new word, their method increases a category's weight if the embedding similarity between the new word and any word associated with the category is greater than some threshold. \citet{onoe2020interpretable} propose a method for interpretable entity representations as a probability vector of entity types. They train text classifiers for each entity type, which is computationally expensive and requires large quantities of training data and labels. Modifying either the predicted entity types or the data domain involves retraining the classifiers.






    

\section{\oursboth}
\label{sec:vanilla}

\begin{algorithm*}
\caption{\oursboth}
\label{algo:vanilla}
\begin{algorithmic}[1]
\State{Initialize deep language model $M_\theta$}
\State{Obtain all categories $C=\{c_i\}_{i=1}^d$ in chosen lexicon $L$}
\If{\ours}
    \State{Obtain all words $S_{c_i}=\{s_{j,c_i}\}_{j=1}^m$ that $L$ maps to category $c_i$}
\ElsIf{\oursp}
    \State{Obtain all sentences $S_{c_i}=\{s_{j,c_i}\}_{j=1}^m$ containing words that $L$ maps to category $c_i$}
\EndIf
\State{$r_{s_{new}} = M_\theta(s_{new})$} \Comment{Get embedding for new sentence $s_{new}$}
\For{$i \in [d]$}
    \State{$\mathcal{R}_{c_i} = \{M_\theta(s_{j,c_i})\}_{j=1}^m$, where $\mathcal{R}_{c_i}=(r_{jk})_{1\leq j \leq m, 1 \leq k \leq n}$} \Comment{Get deep embeddings}
    \For {$k \in [n]$}
        \State{$\mathbf{centroid}(c_i)_k=\frac{1}{m}\sum_{j=1}^m r_{jk}$} \Comment{Get centroid of embeddings}
    \EndFor
    \State{$\mathbf{h}(s_{new})_i = g(r_{s_{new}}, \mathbf{centroid}(c_i))$} \Comment{Compute similarity $g$ of new sentence and centroid}
\EndFor
\State \Return $\mathbf{h}(s_{new})$ \Comment{Return representation of $s_{new}$}
\end{algorithmic}
\end{algorithm*}




\oursboth draws upon the notion of the lexicon; however, rather than mapping words to categories, \oursboth maps the categories to the \textit{deep embeddings of sentences that contain those words}. This, in effect, automatically generates dictionaries of sentence embeddings. To encode the categories of a new sentence, \oursboth uses the similarity between the embedding of the new sentence and the embeddings of the sentences associated with each category.




Generally, \oursboth can be thought of as two parts: (1) building a sentence embedding dictionary and (2) using that dictionary to generate an interpretable representation for a new sentence. We describe the details of the procedure in Sections \ref{sec:building_sentecon} and \ref{sec:generating_sentecon_rep}. The full \oursboth method is formally outlined in Algorithm \ref{algo:vanilla}.

\subsection{Building a sentence embedding lexicon}
\label{sec:building_sentecon}

We present two variants of our approach, \ours and \oursp, both of which are possible ways to build a sentence embedding dictionary. 
An illustration of the two variants can be found Figure \ref{fig:lexicon_to_sentecon}. To begin, suppose we have a traditional lexicon $L$ that maps words to categories.


\textbf{\ours} efficiently approximates sentences for each category using the deep embeddings of the words $L$ associates with that category. Loosely speaking, a word embedding from a language model contains information from all sentences in the training corpus that use that word. As the state-of-the-art pre-trained language models are trained on vast corpora, a word embedding from a pre-trained (or pre-trained, then fine-tuned) language model
will capture in some sense the ``typical'' sentence context for that word. The word embedding can thus be treated as representative of all sentences that use that word. Therefore, the embeddings for all words in a category form a compact representation of all sentences in the training corpus containing any words associated with that category.


\textbf{\oursp} allows our interpretable language representation to further adapt to a particular data domain using only unlabeled text from that domain. Language patterns are not necessarily the same across different domains. Consequently, we can improve how well \ours representations characterize the text in different settings by altering the method by which we construct the sentence embedding dictionary. Specifically, we tailor \ours to the data using a \textit{reference corpus} of unlabeled sentences from the domain of interest. Sentences from the reference corpus are mapped to a category if the sentence contains at least one word that the lexicon $L$ associates with a category.

We use a deep language model $M_\theta$ to produce the embeddings for the words (for \ours) or sentences (for \oursp) $S_{c_i} = \{s_{j,c_i}\}_{j=1}^m$ associated with each category $c_i \in C$, where $C=\{c_i\}^d_{i=1}$. Sentence embeddings are computed via average pooling of token embeddings. This yields a $m \times n$ matrix of embeddings, $\mathcal{R}_{c_i}=\{M_\theta(s_{j,c_i})\}_{j=1}^m$, where $m$ is the number of words or sentences associated with the category and $n$ is the hidden size of $M_\theta$.



\subsection{Generating a \oursboth representation}
\label{sec:generating_sentecon_rep}

After obtaining deep embeddings for all \ours words or \oursp sentences, we find the centroid of the embeddings for each category to obtain a compact and efficient representation of the category.\footnote{If there are thematic or topical groupings of words or sentences within a single category, multiple centroids per category may be used. Therefore, the number of centroids per category can be viewed as a tunable hyperparameter. We elaborate further on this topic in the appendix (Section \ref{sec:centroids}).} For a category $c_i$, the centroid is found by taking the column-wise mean of $\mathcal{R}_{c_i}$, resulting in a $1 \times n$ vector. That is, letting $r_{jk}$ denote the element of $\mathcal{R}_{c_i}$ in row $j$, column $k$, we find the $k$-th element of the centroid as $$\mathbf{centroid}(c_i)_k=\frac{1}{m}\sum_{j=1}^m r_{jk}$$

Given a new sentence $s_{new}$, generating a \oursboth representation requires us to compute the similarity between the new sentence and each of the categories. This is done by first embedding the new sentence as $r_{s_{new}}=M_\theta(s_{new})$, then using a similarity function $g$ to obtain a distance between $r_{s_{new}}$ and each category centroid $\textbf{centroid}(c_i)$. Specifically, for each category $c_i$, $i \in [d]$, we compute the similarity as $g(r_{s_{new}}, \mathbf{centroid}(c_i))$ and assign this value as the weight for category $c_i$. That is, letting $\mathbf{h}(s_{new})$ be the \oursboth representation of $s_{new}$, we have for all $i \in [d]$,
$$\mathbf{h}(s_{new})_i=g(r_{s_{new}}, \mathbf{centroid}(c_i))$$


\section{Experimental Setup}

To assess the utility of \ours and \oursp, we evaluate both methods to determine how well they characterize text in comparison to both existing lexicon-based methods and deep language models. When computing \oursboth representations, we use MPNet \citep{song2020mpnet}
as our deep language model $M_\theta$ and cosine similarity as our similarity metric $g$. Our experiments consist of both human evaluations of \oursboth language representations and tests of performance when using them in downstream predictive tasks.

\subsection{Lexicons}


Linguistic Inquiry and Word Count (\textbf{LIWC}) is a human expert-constructed lexicon generally viewed as a gold standard for lexicons \citep{pennebaker2015development}. Its 2015 version has a vocabulary of 6,548 words that belong to one or more of its 85 categories, most of which are related to psychology and social interaction. We choose to exclude the 33 grammatical categories and retain the remaining 52 topical categories (list in appendix Section \ref{sec:liwc_categories}).

\textbf{Empath} is a semi-automatically generated lexicon with a default vocabulary of 16,159 words that belong to one or more of its 194 categories \citep{fast2016empath}. Empath defines a category using a small number of human-selected seed words, which are used to automatically discover related words that are then also associated with the category. Empath relates words using the cosine similarity of contextualized word embeddings from a deep skip-gram network trained for word prediction, and its categories are chosen from common dependency relationships in the ConceptNet \cite{liu2004conceptnet} knowledge base.

\subsection{Datasets for downstream tasks}
\label{sec:downstream_datasets}

In our performance experiments, we evaluate across several benchmark datasets: Stanford Sentiment Treebank (\textbf{SST}), a collection of polarized sentences from movie reviews \citep{socher2013recursive}; Multimodal EmotionLines Dataset (\textbf{MELD}), a multimodal dialogue dataset from the TV show \textit{Friends} \citep{poria2019meld}; Large Movie Review Dataset (\textbf{IMDb}), which comprises complete movie reviews from the website IMDb \citep{maas-EtAl:2011:ACL-HLT2011}; and Multimodal Opinion-level Sentiment Intensity Corpus (\textbf{MOSI}), a set of opinion video clips from YouTube \citep{zadeh2016mosi}. These datasets were chosen to represent a range of data domains and scenarios in which lexicons like LIWC and Empath would typically be used, such as sentiment analysis, social interaction, and dialogue. Additional details are provided in the appendix (Section \ref{sec:data}).

For each of these datasets, we reserve a held-out set (without labels) to use as the \oursp reference corpus. This allows us to adapt our \oursp representation for the task domain. 



\subsection{Baseline representations and models}
\label{sec:baselines}

Our first evaluation is to compare interpretable representations of sentences with human judgements of those sentences (see Section \ref{sec:human_eval_method}). We have two primary baselines: \textbf{Lexicon} and \textbf{Lexicon+word2vec}. The \textbf{Lexicon} representation uses a bag-of-categories approach to encode the text using a traditional lexicon; in our experiments, we use LIWC and Empath, giving us the lexicon-specific baselines \textbf{Lexicon (L)} and \textbf{Lexicon (E)}, respectively. Bag-of-categories uses a lexicon to label each word in a text with one or more categories. From these categorized words, a vector of category counts can be constructed for a sentence. 

The \textbf{Lexicon+word2vec} language representation implements the previously mentioned soft matching approach proposed by \citet{gong2018improving}. Although the authors describe the method for LIWC only, we generalize the method to Empath also, from which we obtain the baselines \textbf{LIWC+word2vec} and \textbf{Empath+word2vec}. We include this baseline to separate the effects of adding sentence context from the effects of soft matching. In our human evaluation, we focus on LIWC given its broad use in many research areas and use Lexicon (L) and LIWC+word2vec as baselines.

In our downstream prediction experiments, we include an additional baseline model based on recent transformer self-attention architectures, \textbf{MPNet} \cite{song2020mpnet}, to show performance for a non-interpretable language representation. We chose MPNet over other transformer architectures due to its better performance; we report results using other language models in the appendix (Section \ref{sec:sensitivity_lm}). Pre-trained and fine-tuned MPNet are also used as $M_\theta$, the deep language model used to generate sentence embeddings for \oursboth.

Taking both LIWC and Empath as our traditional lexicons, we evaluate \ours and \oursp against Lexicon, Lexicon+word2vec, and MPNet. For all language representations, we add a linear layer over the representation and train the linear layer on the downstream task to obtain our predictions. Details about the training procedures are provided in the appendix (Section \ref{sec:training_details}).

We note that we do not expect \oursboth to outperform non-interpretable transformer-based language models on predictive tasks. We instead view MPNet as a reasonable upper bound for the performance of interpretable approaches.

\subsection{Methodology for human evaluation}
\label{sec:human_eval_method}

As a fair and reliable way to compare \oursboth to other lexicon-based language representations, we collected an extensive set of human sentence-level annotations for all 52 non-grammatical categories of LIWC. In total, 100 sentences randomly sampled from MELD were each annotated across 52 categories by 6 human raters, for a total of 31,200 annotations. These annotations are available as a public dataset on our GitHub repository.

The human annotation study was conducted on the online research platform Prolific.\footnote{\url{https://www.prolific.co/}} To avoid annotator fatigue, the 52 categories were randomly split into 5 sets of roughly equal size, and each set was given its own annotation task. Sentences were annotated in batches of 20, and each annotation task had 6 independent annotators. During the study, each annotator was shown one sentence at a time, alongside one set of 8 to 10 LIWC categories. Annotators were then asked to rate on a scale from 0 to 2 the extent to which each of the categories is expressed. This yielded a human score (averaged over the 6 annotators) of the relevance of each category for each annotated sentence.

We assessed the reliability of our annotations using intraclass correlation coefficients (ICC). Generally speaking, ICC values above 0.50, 0.75, and 0.90 indicate moderate, good, and excellent inter-rater reliability, respectively \citep{koo2016}. We obtained an average ICC estimate of 0.686 with a 95\% confidence interval of [0.606, 0.746], demonstrating moderate to good reliability.

Further details about this study and its results are provided in the appendix (Section \ref{sec:human_evaluation_details}).

\section{Results and Discussion}

\subsection{Human evaluation}

\begin{figure}
    \centering
    \includegraphics[width=0.41\textwidth]{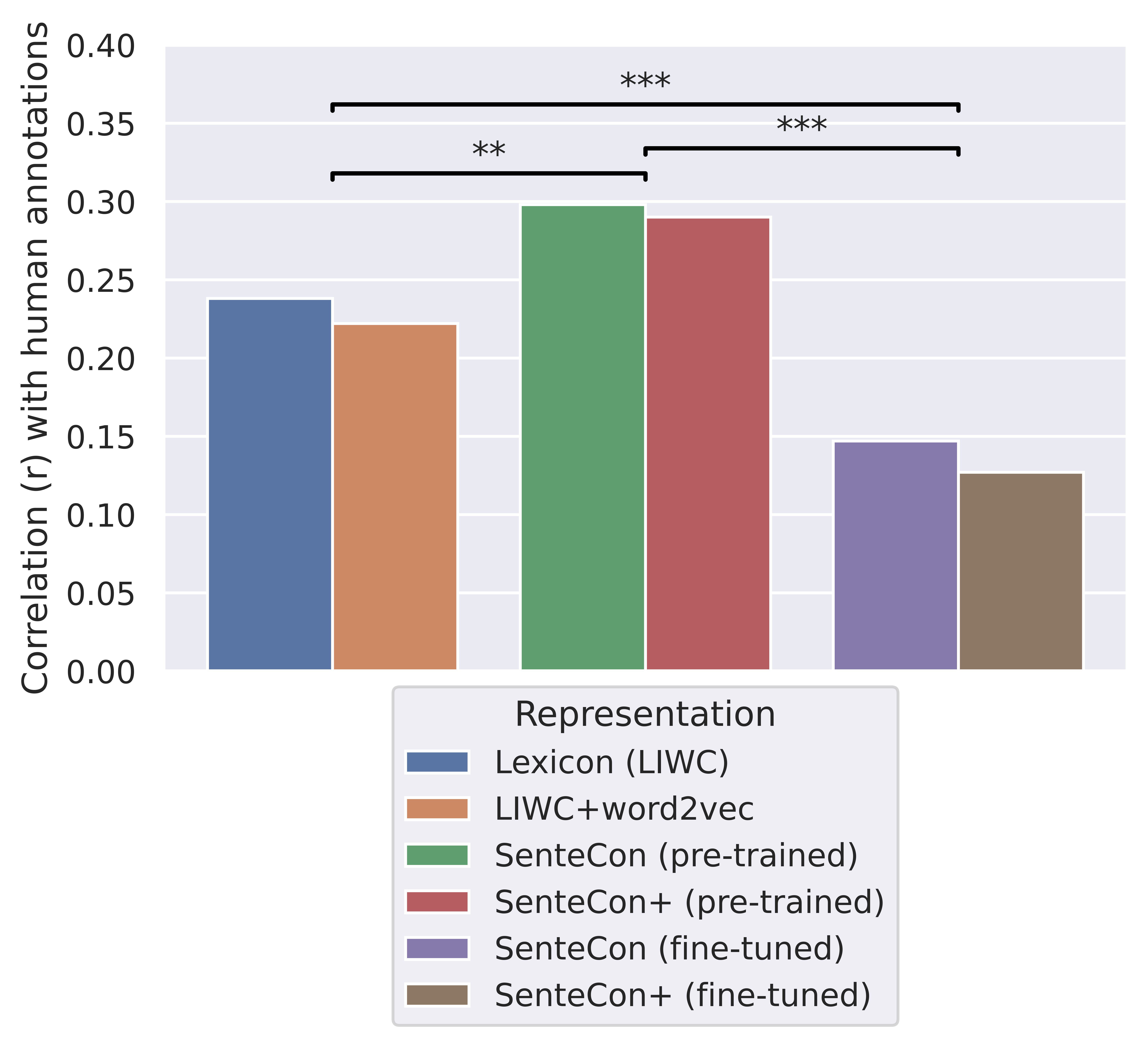}
    \caption{Average Pearson correlations ($r$) between human category annotations and interpretable language representations. $^{**}$ denotes a difference with $p<0.005$, and $^{***}$ denotes a difference with $p<0.0005$.}
    \label{fig:avg_topic_corrs}
\end{figure}

Using the human annotations described in Section \ref{sec:human_eval_method}, we examine how well the different interpretable language representations reflect human perceptions of the text. Across all annotated sentences, we computed Pearson correlations between the human-annotator category scores and the category weights from each sentence representation (Lexicon (L), LIWC+word2vec, \ours with pre-trained $M_\theta$, \oursp with pre-trained $M_\theta$, \ours with $M_\theta$ fine-tuned on MELD, and \oursp with $M_\theta$ fine-tuned on MELD). These results are shown in Figure \ref{fig:avg_topic_corrs}. For illustrative purposes, we include correlations for 10 randomly selected sentences in Table \ref{tab:topic_corrs} in the appendix.

We observe that when $M_\theta$ is pre-trained, \oursboth correlates much more strongly with human category ratings than do either of the existing lexicon methods, Lexicon (L) and LIWC-word2vec. Using a paired two-sided $t$-test, we find that this difference is statistically significant. Importantly, these results suggest that \textbf{when used with a pre-trained $M_\theta$, \ours and \oursp better characterize the text than existing interpretable methods do}, since they are more consistent with human perceptions of the text.

Interestingly, when $M_\theta$ is fine-tuned on the target domain, \oursboth correlates much \textit{less} strongly with human category ratings than the existing lexicon methods do. This difference is also statistically significant. These results suggest that  downstream performance gains from fine-tuning $M_\theta$ may come at a cost to interpretability.

We find no statistically significant difference between \ours and \oursp given the same $M_\theta$. That is, \ours (pre-trained) and \oursp (pre-trained) have no statistically significant difference, nor do \ours (fine-tuned) and \oursp (fine-tuned). We also find no statistically significant difference between Lexicon (L) and LIWC-word2vec.


\begin{table*}[!t]
    \centering
    \begin{tabular}{c|c|c|ccccc}
    \hline
        Representation & Interpretable? &  $M_\theta$ & MELD (e) & MELD (s) & SST & IMDb & MOSI \\
    \hline
        Majority / mean & - & - & 48.1 & 48.1 & 49.9 & 50.0 & -0.001 \\
    \hline
        Lexicon (L) & Yes & - &  46.5 & 49.5 & 67.8 & 76.7 & 0.202 \\
        LIWC+word2vec & Yes & - & 47.5 & 49.4 & 78.7 & 81.4 & 0.270 \\
        \ours (L) & Yes & Pre-trained & 47.7 & 57.6 & 86.5 & 84.2 & \textbf{0.505} \\
        \oursp (L) & Yes & Pre-trained & \textbf{54.6} & \textbf{61.6} & \textbf{88.0} & \textbf{86.3} & 0.487 \\
    \hline
        Lexicon (E) & Yes & - & 39.7 & 44.4 & 63.4 & 74.9 & $\ll 0$ \\
        Empath+word2vec & Yes & - & 46.0 & 50.8 & 81.4 & 85.1 & 0.222 \\
        \ours (E) & Yes & Pre-trained & 51.5 & 59.2 & 88.7 & 87.0 & 0.450 \\
        \oursp (E) & Yes & Pre-trained & \textbf{52.4} & \textbf{60.4} & \textbf{88.9} & \textbf{88.3} & \textbf{0.468} \\
    \hline
        Pre-trained MPNet & No & - & 58.9 & 65.0 & 89.5 & 89.2 & 0.482 \\
    \hline
    \end{tabular}
    \caption{Performance comparisons of \oursboth and traditional lexicon-based methods when used in downstream prediction tasks. (L) indicates that LIWC was used as the base lexicon, while (E) indicates that Empath was used. The best result for each base lexicon choice is bolded. We report test accuracy for MELD (on both emotion and sentiment tasks), SST, and IMDb and test $R^2$ for MOSI.}
    \label{tab:sentecon_results}
\end{table*}

\begin{table*}[t]
    \centering
    \begin{tabular}{c|c|c|ccccc}
    \hline
        Representation & Interpretable? &  $M_\theta$ & MELD (e) & MELD (s) & SST & IMDb & MOSI \\
    \hline
        \ours (L) & Yes & Fine-tuned & 57.2 & 68.1 & \textbf{93.4} & \textbf{95.1} & 0.672 \\
        \oursp (L) & Yes & Fine-tuned & \textbf{59.9} & \textbf{68.1} & 93.2 & 95.0 & \textbf{0.673} \\
    \hline
        \ours (E) & Yes & Fine-tuned & 56.3 & 67.3 & 93.2 & 94.9 & \textbf{0.709} \\
        \oursp (E) & Yes & Fine-tuned & \textbf{59.3} & \textbf{68.5} & \textbf{93.3} & \textbf{95.0} & 0.702 \\
    \hline
        Fine-tuned MPNet & No & - & 59.8 & 67.8 & 93.4 & 95.1 & 0.694 \\
    \hline
    \end{tabular}
    \caption{Performance comparisons of  \oursboth and deep language representations when used in downstream prediction tasks. (L) indicates that LIWC was used as the base lexicon, while (E) indicates that Empath was used. The best result for each base lexicon choice is bolded. We report test evaluation metrics.}
    \label{tab:sentecon_ft_results}
\end{table*}

\subsection{Performance on downstream tasks}
\label{sec:perf_downstream}

We evaluate the implications of \oursboth on downstream predictive performance. Our results, including comparisons with baseline models, are shown in Tables \ref{tab:sentecon_results} and \ref{tab:sentecon_ft_results}. Importantly, we find that:

\textbf{(1) Both \ours and \oursp perform better than the Lexicon and Lexicon+word2vec approaches do on downstream tasks} (Table \ref{tab:sentecon_results}). This finding suggests that by modeling sentence-level context, \ours and \oursp improve text characterization with respect to not only human evaluation but also downstream prediction. Across all classification tasks (MELD, SST, and IMDb), \ours and \oursp achieve substantially higher accuracy than Lexicon and Lexicon+word2vec do, regardless of whether LIWC or Empath is used as the base lexicon. Likewise, \ours and \oursp achieve substantially higher $R^2$ on the MOSI regression task than Lexicon and Lexicon+word2vec do.

\textbf{(2) When used with a fine-tuned $M_\theta$, \ours and \oursp provide interpretability to deep language models at no cost to performance} (Table \ref{tab:sentecon_ft_results}). Across all downstream tasks, \oursboth representations---particularly \oursp representations---with fine-tuned $M_\theta$ achieve virtually equal performance compared to fine-tuned MPNet, the deep language model over which they are constructed. This observation holds for both choices of base lexicon $L$. We must emphasize the significance of this result: we are able to construct a layer of high-level interpretable concepts, pass it into a single linear layer (itself an interpretable model), and predict a target with equal performance as if we had used a non-interpretable deep language model fine-tuned on the task. In other words, \textit{we can clearly understand the relationship between these interpretable concepts and the target without compromising performance}. This type of interpretability is far beyond that achieved by existing analyses of deep language models, and this type of performance is far beyond that achieved by existing lexicon-based methods.


\begin{table*}[!h]
    \centering
    \begin{tabular}{c|c|c|cccc}
    \hline
        Word & Meaning 1 & Meaning 2 & \makecell{Matching-sense \\ similarity} & \makecell{Opposing-sense \\ similarity} & \makecell{Individual \\ similarity ratio} \\
    \hline
    \textit{bright} & \textit{shining} & \textit{intelligent} & \textbf{0.692} & 0.608 & 1.139$^{**}$ \\
    \textit{hard} & \textit{forceful} & \textit{difficult} & \textbf{0.677} & 0.539 & 1.256$^{***}$ \\
    \textit{dull} & \textit{boring} & \textit{unintelligent} & \textbf{0.686} & 0.591 & 1.161$^{***}$ \\
    \textit{dark} & \textit{dim} & \textit{sinister} & \textbf{0.614} & 0.488 & 1.258$^{***}$ \\
    \textit{cool} & \textit{calm} & \textit{impressive} & \textbf{0.419} & 0.292 & 1.433$^{***}$ \\
    \hline
    \end{tabular}
    \caption{Similarities between contextualized \ours representations of homonyms and their matching- and opposing-sense meanings. $^{**}$ denotes a difference with $p<0.005$, and $^{***}$ denotes a difference with $p<0.0005$.}
    \label{tab:word_sense}
\end{table*}

\begin{figure*}[!h]
    \centering
    \includegraphics[width=0.29\textwidth]{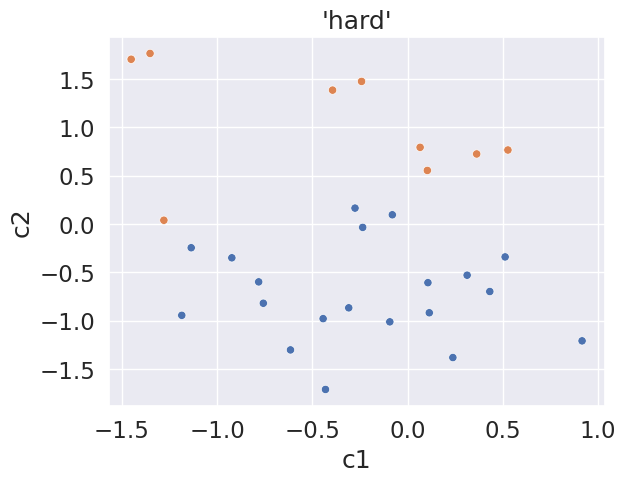}
    \includegraphics[width=0.29\textwidth]{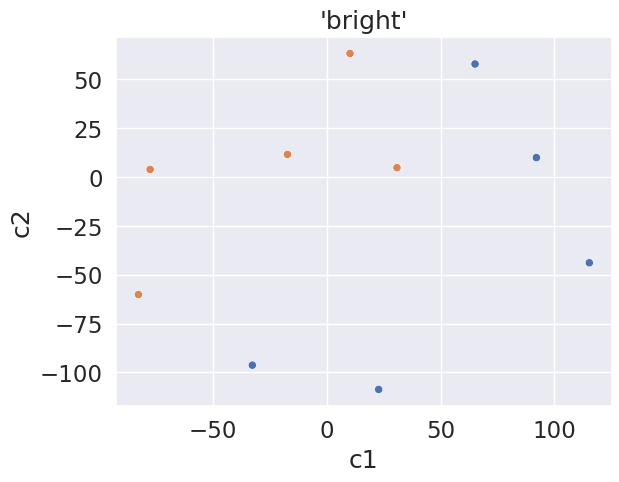} \hspace{0.07cm}
    \includegraphics[width=0.38\textwidth]{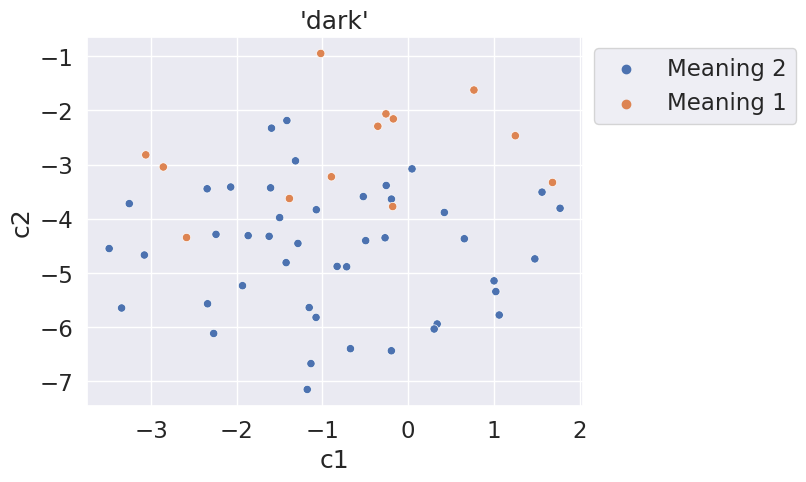}
    \caption{t-SNE plots of contextualized \ours representations of homonyms show separation by word sense.}
    \label{fig:word_sense}
\end{figure*}

\textbf{(3) \oursp offers performance improvements over \ours without negatively impacting interpretability} (Tables \ref{tab:sentecon_results} and \ref{tab:sentecon_ft_results}), supporting the utility of using a reference corpus from the task data domain to refine \ours representations. While fine-tuning $M_\theta$ allows \oursboth to achieve the best performance, it does so at some cost to how well the representation agrees with human evaluations (Figure \ref{fig:avg_topic_corrs}). When human agreement is a priority---e.g., in applications like healthcare and psychology---it may be more desirable to use \oursp with a pre-trained $M_\theta$ instead. This configuration confers performance gains over \ours without compromising human agreement. Furthermore, even when $M_\theta$ is fine-tuned, \oursp still often outperforms \ours, particularly when Empath is the base lexicon $L$.

\subsection{Model analysis: Word sense}

Given these results, we would like to gain some understanding of how \oursboth is able to improve on existing lexicon-based interpretable language representations. 

Prior work on BERT has demonstrated that its strength as a language representation lies partially in its ability to distinguish different word senses based on sentence context \citep{reif2019visualizing, wiedemann2019does, schmidt2020bert}. We postulate that sentence context similarly enables \oursboth to distinguish different word senses, yielding the observed empirical gains in interpretability and performance. To explore this hypothesis, we conduct an experiment to verify whether a word’s sentence context changes its \ours representation to be more similar to its true meaning in the sentence.

\subsubsection{Method}

\textbf{Collecting homonyms.} We first selected words with multiple common meanings (homonyms)---for example, the word \textit{bright}. We began with a list of homonyms compiled from online sources.\footnote{\url{https://7esl.com/homonyms/}}$^{,}$\footnote{\url{https://examples.yourdictionary.com/examples-of-homonyms.html}} For each homonym on the list, we collected all sentences in MELD and SST containing the word. We chose the dataset with more sentences containing the word, and we retained all homonyms for which there were 10 or more associated sentences. We annotated each sentence with the word's corresponding meaning (e.g., we labeled the sentences as using \textit{bright} either to mean \textit{shining} or to mean \textit{intelligent}). For every sentence, this yields a ``matching-sense'' meaning and an ``opposing-sense'' meaning. We retained all homonyms for which each meaning of the word had 5 or more associated sentences.

\textbf{Distinguishing word sense.} With this set of homonyms, we verified whether \ours is capable of distinguishing word sense using a procedure similar to one in \citet{reif2019visualizing} for BERT representations. For each sentence, we obtained the contextualized \ours representation for the selected homonym. We also obtained the non-contextualized \ours representations of three keywords for each meaning of the word (e.g., for \textit{bright}, these keywords are (1) \textit{shining}, \textit{vivid}, \textit{beaming} and (2) \textit{intelligent}, \textit{smart}, \textit{clever}). These keywords were randomly selected from the Oxford English Dictionary synonyms for each meaning of the word. Then---again for each sentence---we computed the cosine similarity between the \ours representations of the homonym and its matching-sense keywords, then the similarity between the \ours representations of the homonym and its opposing-sense keywords.

\subsubsection{Results}

The results of this experiment, which we report in Table \ref{tab:word_sense}, indicate that \ours representations are indeed able to distinguish different word senses. When used in a particular sentence context, words with multiple meanings show significantly more similarity to their matching-sense definition than they do to their opposing-sense definition. We formalize this with the \textit{individual similarity ratio} metric defined by \citet{reif2019visualizing}, which is the ratio of matching-sense similarity to opposing-sense similarity. If a representation is able to correctly distinguish word sense, this ratio should be greater than 1, which we observe to be the case across all selected homonyms. Additionally, $t$-tests indicate that the difference in similarity is statistically significant across all homonyms. 

We further visualize the separation of word senses via t-SNE plots of our \ours representations, similar to experiments by \citet{wiedemann2019does} on BERT embeddings. These plots show that \ours representations of the same word separate clearly in embedding space according to their meanings (Figure \ref{fig:word_sense}).

These results support our claim that \oursboth uses sentence context to improve interpretability and performance on downstream tasks. The ability to distinguish word senses helps \oursboth to correctly identify relevant categories where traditional lexicons may be not be able to do so, thereby allowing \oursboth to better characterize the text.

\section{Conclusion}

In this paper we introduced \ours, a human-interpretable language representation that captures sentence context while retaining the benefits of interpretable lexicons. We conducted human evaluations to determine the agreement between \ours representations and the actual content of the text, and we ran a series of experiments using \ours in downstream predictive tasks. In doing so, we demonstrated that \ours and its extension, \oursp, better represent the character and content of the text than traditional lexicons do. Furthermore, we showed that when used in conjunction with language models fine-tuned on the downstream task, \ours and \oursp provide interpretability to deep language models without any loss of performance. These findings render \ours and \oursp compelling candidates for problems in fields like medicine, social science, and psychology, where understanding language use is an important part of the scientific process and where insight into a model's decision-making process can be paramount.


\section{Acknowledgements}

This material is based upon work partially supported by National Science Foundation awards 1722822 and 1750439, and National Institutes of Health awards R01MH125740, R01MH132225, R01MH096951 and R21MH130767. Any opinions, findings, conclusions, or recommendations expressed in this material are those of the author(s) and do not necessarily reflect the views of the sponsors, and no official endorsement should be inferred.

\section{Limitations}


We recognize that several limitations remain with \ours and \oursp. 

(1) Despite the gains in performance obtained by using a fine-tuned $M_\theta$ with \ours, we note that this version of \ours has significantly worse agreement with human evaluation than when a pre-trained $M_\theta$ is used. It is not immediately obvious why this should be the case. Although it is always possible to use \oursp with a pre-trained $M_\theta$ in cases where agreement with human evaluation is particularly important, future work should examine why this degradation occurs and explore whether it is possible to maintain human agreement while also seeing those same performance gains (possibly through a secondary loss term that prioritizes human agreement).

(2) When building a sentence embedding dictionary, the base lexicon of \oursboth may map lexically similar sentences to the same categories, regardless of attributes like negation. Despite this, \ours produces meaningful representations for sentences that require compositional understanding, which we attribute to the large number of sentences mapped to each category (recall that each contextualized word embedding mapped to a category can be viewed as a summary of all sentences in the language model pre-training corpus containing that word). For example, the number of negated sentences in the sentence embedding dictionary is far smaller than the number of non-negated sentences---and likewise for other attributes requiring compositional parsing. Consequently, each category's centroid is still approximately an average of the non-negated sentences.

The same principle applies to \oursp if a reasonably-sized reference corpus is used. If, however, only a very small reference corpus is available and the task dataset is known to require strong compositional understanding, \ours should be used instead of \oursp.


\section{Ethics Statement}

\textbf{Broader impact.} As deep language models gain greater prominence in both research and real-world use cases, concerns have arisen regarding their opaque nature \cite{rudin2019stop, arrieta2020explainable}, their tendency to perpetuate and even amplify social biases in the data on which they are trained \cite{bolukbasi2016man, swinger2019biases, caliskan2017semantics}, and their encoding of spurious relationships between the target and irrelevant parts of the input \cite{veitch2021counterfactual}. Particularly given their increasing deployment in healthcare, psychology, and social science, as we mention earlier in this paper, it is crucial that these black-box models be rendered more transparent to ensure that decisions are being made in a principled way. In other words, interpretability is not only an intellectual goal but also an ethical one.

In service of this goal, our proposed language representation, \ours, provides clear insight into the relationship between human-interpretable concepts and outcomes of interest in machine learning tasks. It is able to do so without negatively impacting predictive performance---an important factor, since a primary motivator for using non-interpretable language representations is their excellent performance on machine learning tasks. We hope that this will motivate others to use \ours, and we also hope that using \ours will allow users to better understand how their machine learning pipelines make decisions, evaluate their models for bias, and enforce correct and robust relationships between inputs and outputs.

\textbf{Ethical considerations.} This work involves the collection of new data to assess the consistency of \oursboth representations with human annotations of the content of text passages. No information was collected about the annotators, and the data is not sensitive in nature. In the course of data collection, we took measures to ensure fair compensation and treatment of annotators. Annotators were provided a description of the study and given the option to decline the study after learning its details, and all annotators were paid at a rate above the local minimum wage.

\oursboth relies on pre-trained deep language models to compute language representations. Our use of these pre-trained models is limited to research purposes only and is compliant with their intended use. We acknowledge that the use of pre-trained models introduces the possibility that \oursboth may encode some biases contained in those models. As a consequence, interpretations of the relationships between \oursboth categories and targets (when using \oursboth in modeling) may also contain elements of bias.

\bibliography{ref}
\bibliographystyle{acl_natbib}

\appendix

\section{Effects of \oursboth parameter choices}
\label{sec:sensitivity}

To ensure that our findings in Section \ref{sec:perf_downstream} are robust to different parameter choices in \oursboth, we conduct analyses over the number of centroids per category, choice of deep language model $M_\theta$, and choice of reference corpus. We take LIWC as our base lexicon for all experiments.

\subsection{Number of centroids per category}
\label{sec:centroids}

\begin{table*}[!ht]
    \centering
    \begin{tabular}{c|c|cccccc}
    \hline
        Representation & \# centroids & MELD (e) & MELD (s) & SST & IMDb & MOSI \\
    \hline
        \ours (L) & 1 & \textbf{57.2} & \textbf{68.1} & 93.4 & \textbf{95.1} & 67.2 \\
        \ours (L) & 2 & 55.8 & 67.7 & 93.1 & 94.9 & 67.8 \\
        \ours (L) & 3 & 55.6 & 67.4 & 93.5 & 94.9 & \textbf{69.3} \\
        \ours (L) & 4 & 55.5 & 67.2 & \textbf{93.5} & \textbf{95.1} & 68.4 \\
    \hline
        \oursp (L) & 1 & \textbf{59.9} & \textbf{68.1} & \textbf{93.2} & \textbf{95.0} & \textbf{67.3} \\
        \oursp (L) & 2 & 59.0 & 67.4 & 92.8 & 94.7 & 66.9 \\
        \oursp (L) & 3 & 57.6 & 68.0 & 93.2 & 93.6 & - \\
        \oursp (L) & 4 & 55.3 & 66.9 & 93.1 & 93.8 & - \\
    \hline
    \end{tabular}
    \caption{Performance comparisons of \oursboth across different numbers of centroids per category. We use LIWC as the base lexicon and fine-tuned MPNet as $M_\theta$. We report test accuracy for MELD, SST, and IMDb and test $R^2$ for MOSI.}
    \label{tab:num_centroids}
\end{table*}

If our lexicon categories are very broad, we may have reason to believe that it would be useful to have multiple centroids per category, rather than summarizing the category as a single centroid. Here, we report the effects of different numbers of centroids per category on \oursboth performance on downstream tasks.

To define multiple centroids for a given category, we use an unsupervised clustering method to create $P$ clusters of word or sentence embeddings for each category. For each of the $P$ clusters, we compute the centroid as before, so we now have $P$ centroids for every category.

Now, given a new sentence $s_{new}$, we compute the similarity between the new sentence and \textit{each centroid of each category}. Then when computing our \oursboth representation, the weight for category $c_i$ is taken to be the largest similarity between $s_{new}$ and any one of the centroids for $c_i$. That is, letting $\mathbf{centroid}(c_i)_p$ be the $p$-th cluster centroid for category $c_i$ and $\mathbf{h}(s_{new})_i$ again be the \oursboth weight for $c_i$,
$$\mathbf{h}(s_{new})_i = \max_{p \in P}(g(r_{s_{new}}, \mathbf{centroid}(c_i)_p)$$

Across our evaluation tasks, we do not find additional centroids to produce substantial performance gains (Table \ref{tab:num_centroids}), though small improvements are observed for \ours on SST and MOSI. We encourage users of \oursboth to treat the number of centroids as a tunable hyperparameter---but in many cases, including the ones we explore in our experiments, a single centroid per category should be sufficient.

\subsection{Choice of language model}
\label{sec:sensitivity_lm}

Here, we report the effects of different choices of $M_\theta$ model architectures on \ours performance on downstream tasks. All language models are pre-trained.

\begin{table*}[!ht]
    \centering
    \begin{tabular}{c|c|cccccc}
    \hline
        Representation & $M_\theta$ & MELD (e) & MELD (s) & SST & IMDb & MOSI \\
    \hline
        Lexicon (L) & - & 46.5 & 49.5 & 67.8 & 76.7 & 0.202 \\
        LIWC+word2vec & - & 47.5 & 49.4 & 78.7 & 81.4 & 0.270 \\
    \hline
        \ours (L) & MPNet & 47.7 & 57.6 & \textbf{86.5} & \textbf{84.2} & \textbf{0.505} \\
        \ours (L) & MiniLM & 50.7 & 56.4 & 77.9 & 75.7 & 0.411 \\
        \ours (L) & DistilRoBERTa & 48.6 & 54.7 & 85.2 & 82.4 & 0.289 \\
        \ours (L) & BERT & 58.7 & \textbf{65.4} & 81.3 & 84.4 & 0.364 \\
        \ours (L) & RoBERTa & \textbf{56.5} & 60.7 & 79.4 & 83.7 & 0.118 \\
    \hline
        Pre-trained embedding & MPNet & 58.9 & 65.0 & 89.5 & 89.2 & \textbf{0.482} \\
        Pre-trained embedding & MiniLM & 59.9 & 64.7 & 81.3 & 81.1 & 0.150 \\
        Pre-trained embedding & DistilRoBERTa & 58.5 & 64.9 & 88.3 & 87.6 & 0.264 \\
        Pre-trained embedding & BERT & 56.8 & 63.2 & 86.1 & 89.1 & 0.259 \\
        Pre-trained embedding & RoBERTa & \textbf{60.5} & \textbf{65.0} & \textbf{90.3} & \textbf{92.0} & 0.177 \\
    \hline
    \end{tabular}
    \caption{Performance comparisons of \ours when used with different pre-trained language models as $M_\theta$ in downstream prediction tasks. We report test accuracy for MELD, SST, and IMDb and test $R^2$ for MOSI.}
    \label{tab:sensitivity_lm}
\end{table*}

To determine the impact of selecting a well-performing language model as our $M_\theta$, we construct additional \ours representations using pre-trained DistilRoBERTa \cite{sanh2020distilbert}, MiniLM \cite{wang2020minilm}, BERT \cite{devlin2019bert}, and RoBERTa \cite{liu2019roberta}, all of which are transformer-based language models like MPNet. Comparing the performance of \ours representations to Lexicon and Lexicon-word2vec, \textit{we observe that \ours continues to outperform both baselines across all choices of $M_\theta$} (Table \ref{tab:sensitivity_lm}), even for the smaller MiniLM model. 

\ours performance seems to scale generally---though not perfectly---with $M_\theta$ performance. For example, MPNet and RoBERTa are the best-performing pre-trained language models, and \ours with MPNet and RoBERTa as $M_\theta$ are the best-performing variants of \ours (aside from the MELD sentiment task, where \ours with BERT achieves the best performance).

\subsection{Choice of reference corpus}
\label{sec:sensitivity_ref}

\begin{table}[!ht]
    \centering
    \begin{tabular}{c|cc}
    \hline
        Reference corpus & MELD (e) & MELD (s) \\
    \hline
        None & 50.7 & 56.4 \\
        \hline
        MELD & \textbf{55.5} & \textbf{61.3} \\
        Switchboard & 49.7 & 55.6 \\
        NYT & 49.9 & 53.9 \\
        PubMed & 50.6 & 55.7 \\
    \hline
    \end{tabular}
    \caption{Performance comparisons of \oursp on MELD when used with different reference corpora. We use LIWC as the base lexicon and pre-trained MiniLM as $M_\theta$, and we report test accuracies.}
    \label{tab:sensitivity_refcorp}
\end{table}

In Section \ref{sec:downstream_datasets}, we describe our approach for creating a reference corpus: using a held-out portion of the task dataset. However, it is useful to know whether the reference corpus must be from the \textit{same} domain as the task or whether a reference corpus from a \textit{similar} domain may suffice to improve performance over \ours. With MELD as our downstream task dataset, we select as our reference corpora one dataset that is similar to MELD (\textbf{Switchboard}, a series of utterances from dyadic phone conversations); one that is moderately different (\textbf{NYT}\footnote{\url{https://www.kaggle.com/datasets/benjaminawd/new-york-times-articles-comments-2020}}, a dataset of New York Times article summaries from 2020); and one that is extremely different (\textbf{PubMed}, a collection of abstracts from academic papers published in medical journals) \citep{godfrey1992switchboard, dernoncourt2017pubmed}. More details about these datasets are provided in Section \ref{sec:data}. To reduce computational load, we use the smaller transformer-based language model MiniLM \citep{wang2020minilm} as our $M_\theta$.

We evaluate \oursp representations on the MELD emotion and sentiment classification tasks using the three new reference corpora (Table \ref{tab:sensitivity_refcorp}). We find that 
using any of the three new reference corpora yields worse performance than using a held-out set from MELD (and in fact, worse performance than not using a reference corpus at all). These results support the conclusion that the reference corpus should be from the \textit{same} domain as the task. Only \oursp with a reference corpus consisting of a portion of the task dataset itself provides performance improvements over \ours with no reference corpus.

\section{Experimental Details}

\subsection{LIWC categories}
\label{sec:liwc_categories}

The full list of non-grammatical LIWC categories used in our experiments is as follows: \textit{affect}, \textit{posemo}, \textit{negemo}, \textit{anx}, \textit{anger}, \textit{sad}, \textit{social}, \textit{family}, \textit{friend}, \textit{female}, \textit{male}, \textit{cogproc}, \textit{insight}, \textit{cause}, \textit{discrep}, \textit{tentat}, \textit{certain}, \textit{differ}, \textit{percept}, \textit{see}, \textit{hear}, \textit{feel}, \textit{bio}, \textit{body}, \textit{health}, \textit{sexual}, \textit{ingest}, \textit{drives}, \textit{affiliation}, \textit{achiev}, \textit{power}, \textit{reward}, \textit{risk}, \textit{focuspast}, \textit{focuspresent}, \textit{focusfuture}, \textit{relativ}, \textit{motion}, \textit{space}, \textit{time}, \textit{work}, \textit{leisure}, \textit{home}, \textit{money}, \textit{relig}, \textit{death}, \textit{informal}, \textit{swear}, \textit{netspeak}, \textit{assent}, \textit{nonflu}, \textit{filler}.

The list of excluded grammatical LIWC categories is as follows: \textit{function}, \textit{pronoun}, \textit{ppron}, \textit{i}, \textit{we}, \textit{you}, \textit{shehe}, \textit{they}, \textit{ipron}, \textit{article}, \textit{prep}, \textit{auxverb}, \textit{adverb}, \textit{conj}, \textit{negate}, \textit{verb}, \textit{adj}, \textit{compare}, \textit{interrog}, \textit{number}, \textit{quant}.

\subsection{Empath categories}

The full list of Empath categories used in our experiments is as follows: 
\textit{help, office, dance, money, wedding, domestic\_work, sleep, medical\_emergency, cold, hate, cheerfulness, aggression, occupation, envy, anticipation, family, vacation, crime, attractive, masculine, prison, health, pride, dispute, nervousness, government, weakness, horror, swearing\_terms, leisure, suffering, royalty, wealthy, tourism, furniture, school, magic, beach, journalism, morning, banking, social\_media, exercise, night, kill, blue\_collar\_job, art, ridicule, play, computer, college, optimism, stealing, real\_estate, home, divine, sexual, fear, irritability, superhero, business, driving, pet, childish, cooking, exasperation, religion, hipster, internet, surprise, reading, worship, leader, independence, movement, body, noise, eating, medieval, zest, confusion, water, sports, death, healing, legend, heroic, celebration, restaurant, violence, programming, dominant\_heirarchical, military, neglect, swimming, exotic, love, hiking, communication, hearing, order, sympathy, hygiene, weather, anonymity, trust, ancient, deception, fabric, air\_travel, fight, dominant\_personality, music, vehicle, politeness, toy, farming, meeting, war, speaking, listen, urban, shopping, disgust, fire, tool, phone, gain, sound, injury, sailing, rage, science, work, appearance, valuable, warmth, youth, sadness, fun, emotional, joy, affection, traveling, fashion, ugliness, lust, shame, torment, economics, anger, politics, ship, clothing, car, strength, technology, breaking, shape\_and\_size, power, white\_collar\_job, animal, party, terrorism, smell, disappointment, poor, plant, pain, beauty, timidity, philosophy, negotiate, negative\_emotion, cleaning, messaging, competing, law, friends, payment, achievement, alcohol, liquid, feminine, weapon, children, monster, ocean, giving, contentment, writing, rural, positive\_emotion, musical}.


\subsection{Human evaluation study details}
\label{sec:human_evaluation_details}

\textbf{Question.} In the human evaluation study, annotators were asked the following question:

\textit{For each of the following topics or categories, please rate to what extent the topic is expressed in the language, content, and meaning of the sentence. It is possible that none of the topics may be expressed; it is also possible that the topic you feel is most strongly expressed is not present.}

\textit{If a topic is marked with an asterisk, please hover your cursor over each topic for a more detailed description of the topic.}

They were asked to rate according to the following scale and were provided with the accompanying descriptions.
\begin{itemize}
    \item \textit{Not expressed}: Out of all possible interpretations of the sentence above, you cannot imagine a scenario in which the speaker of the sentence was expressing the topic.

    \item \textit{Potentially expressed}: You can imagine at least one scenario in which the speaker of the sentence was expressing the topic.

    \item \textit{Most likely expressed}: The most natural interpretation of the sentence clearly expresses the topic.
\end{itemize}

\textbf{Category batches.} As mentioned in the main paper, the 52 LIWC categories were randomly split into 5 sets of roughly equal size to avoid annotator fatigue. The splits were as follows:

\begin{itemize}
    \item Batch 1: \textit{netspeak}, \textit{differ}, \textit{cause}, \textit{nonflu}, \textit{discrep}, \textit{drivers}, \textit{relig}, \textit{swear}, \textit{feel}, \textit{home}, \textit{family}
    \item Batch 2: \textit{leisure}, \textit{sexual}, \textit{see}, \textit{bio}, \textit{certain}, \textit{money}, \textit{percept}, \textit{female}, \textit{death}, \textit{anger}, \textit{cogproc}
    \item Batch 3: \textit{filler}, \textit{sad}, \textit{posemo}, \textit{friend}, \textit{relativ}, \textit{ingest}, \textit{body}, \textit{work}, \textit{time}, \textit{social}, \textit{informal}
    \item Batch 4: \textit{focusfuture}, \textit{anx}, \textit{affiliation}, \textit{motion}, \textit{power}, \textit{reward}, \textit{space}, \textit{tentat}, \textit{risk}, \textit{focuspresent}, \textit{affect}
    \item Batch 5: \textit{negemo}, \textit{hear}, \textit{male}, \textit{health}, \textit{insight}, \textit{achiev}, \textit{focuspast}, \textit{assent}
\end{itemize}

\textbf{Inter-rater reliability.} To assess the reliability of our annotations, we calculated intraclass correlation coefficients (ICCs) using the \textit{agreement} software package \citep{girard2020}. For each batch of sentences, we computed the ICC and its 95\% confidence interval, then averaged these across category batches (Table \ref{tab:icc}). We averaged ICCs over all batches to obtain the overall ICC.

\begin{table}
    \centering
    \begin{tabular}{c|c}
        \hline
        Category batch & ICC \\
        \hline
        1 & \textbf{0.580} [0.467, 0.662] \\
        2 & \textbf{0.688} [0.603, 0.749] \\
        3 & \textbf{0.730} [0.669, 0.777] \\
        4 & \textbf{0.715} [0.654, 0.763] \\
        5 & \textbf{0.718} [0.635, 0.777] \\
        \hline
        Average & \textbf{0.686} [0.606, 0.746] \\
        \hline
    \end{tabular}
    \caption{ICCs of human annotations of sentence categories across category batches, with 95\% confidence intervals.}
    \label{tab:icc}
\end{table}

\textbf{Annotators.} Annotators were required to be fluent in English and to be nationals of one of the following countries: the United States, the United Kingdom, Ireland, Australia, or Canada.

Annotators were further required to have a prior approval rating of $\geq$ 95\%, and an attention check question was included in every sentence batch. All annotators passed the attention check.

We took care to compensate annotators at a rate above the local minimum wage. Annotators received an average hourly wage of 8.00 USD.

\begin{table*}[!ht]
    \centering
    \begin{tabular}{c|cccc|c}
    \hline
        Dataset & $\text{n}_{\text{train}}$ & $\text{n}_{\text{test}}$ & $\text{n}_{\text{reference}}$ & $\text{n}_{\text{total}}$ & License \\
    \hline
        MELD & 9,989 & 2,610 & 1,109 & 13,708 & GPL-3.0 \\
        SST & 6,920 & 1,821 & 872 & 9,613 & Unknown \\
        IMDb & 25,000 & 15,000 & 10,000 & 50,000 & Unknown \\
        MOSI & 1,034 & 500 & 665 & 2,199 & Other\footnotemark \\
        Switchboard & - & - & 15,000 & - & Other\footnotemark\\
        NYT & - & - & 16,784 & - & CC BY-NC-SA 4.0 \\
        PubMed & - & - & 15,000 & - & Unknown \\
    \hline
    \end{tabular}
    \caption{Composition of dataset splits. The number of train, test, and reference corpus samples is given, along with total samples for each dataset. Licensing information is also given.}
    \label{tab:data_splits}
\end{table*}

\subsection{Data}
\label{sec:data}

Details of train, test, and reference corpus splits are provided in Table \ref{tab:data_splits}, including dataset composition and licensing information. For datasets released with existing train and test splits, we split the existing test set into a reference corpus and new test set. As mentioned in the main paper, all datasets are already publicly available, and the additional splits created for the reference corpora are available on our GitHub repository. All datasets are in English.

\footnotetext[6]{\url{https://github.com/A2Zadeh/CMU-MultimodalSDK/blob/master/LICENSE.txt}}
\footnotetext[7]{\url{https://catalog.ldc.upenn.edu/license/ldc-non-members-agreement.pdf}}

\subsection{Training details}
\label{sec:training_details}

Our language models were built on the HuggingFace\footnote{\url{https://huggingface.co/}} \verb|transformers| library (version 4.16.2), with pre-trained models taken from the HuggingFace model hub. When fine-tuning these models on the task datasets, we used an Adam optimizer and learning rates [$10^{-1}, 10^{-2}, 10^{-3}, 10^{-4}, 10^{-5}$], and we found $10^{-5}$ to be the best learning rate across all models. We trained for 15 epochs and selected the model with the best 5-fold cross-validation loss. All other hyperparameters were set to Trainer class defaults from the \verb|transformers| library.

The number of parameters for each of the deep language models used is reported in Table \ref{tab:num_params}. The license names for the models are also provided.

\begin{table*}[!ht]
    \centering
    \begin{tabular}{c|c|c|c}
    \hline
        Language model & \# dimensions & \# parameters & License \\
    \hline
        MPNet & 768 & 109,486,464 & MIT \\
        RoBERTa & 768 & 124,645,632 & MIT \\
        BERT & 768 & 109,482,240 & Apache-2.0 \\
        MiniLM & 384 & 22,713,216 & Apache-2.0 \\ 
        DistilRoBERTa & 768 & 82,118,400 & Apache-2.0 \\
    \hline
    \end{tabular}
    \caption{Number of dimensions, parameters, and license for each deep language model.}
    \label{tab:num_params}
\end{table*}


\subsection{Computing resources}

\oursboth requires only using an existing deep language model to generate embeddings and consequently is not particularly computationally demanding. Fine-tuning deep language models is more resource-intensive, but we use these only to a limited extent in our experiments, and only on small datasets. We estimate the number of GPU hours used in these experiments to be around 20. All experiments were conducted on machines with consumer-level NVIDIA graphics cards.

\begin{table*}[!ht]
    \centering
    \begin{tabular}{p{2.3cm}|cccccc}
        \hline
        Sentence & \makecell{Lexicon\\ (L)} & \makecell{LIWC+\\ word2vec} & \makecell{\ours\\ (pre-trained)} & \makecell{\oursp\\ (pre-trained)} & \makecell{\ours\\ (fine-tuned)} &
        \makecell{\oursp\\ (fine-tuned)} \\
        \hline
        \textit{What?} & 0.112 & 0.218 & -0.054
        & \textbf{0.251} & 0.223 & -0.129 \\
        \hline
        \textit{Really?} & \textbf{0.211} & -0.153 & 0.175 & -0.011 & 0.147 & 0.089 \\
        \hline
        \textit{It's really sweet and---and tender.} & 0.001 & 0.284 & 0.273 & \textbf{0.325} & 0.260 & 0.003 \\
        \hline
        \textit{Tell her to wear her own earrings.} & 0.222 & 0.239 & 0.307 & \textbf{0.445} & 0.260 & 0.003 \\
        \hline
        \textit{This is totally your fault!} & 0.453 & 0.358 & 0.663 & \textbf{0.672} & 0.465 & 0.409 \\
        \hline
        \textit{My first time with Carol was...} & 0.166 & 0.234 & 0.456 & \textbf{0.487} & -0.041 & 0.126 \\
        \hline
        \textit{No! Ah-ah-ah-ah-ah! You can have this back when the five pages are done! Ahh!} & -0.064 & \textbf{0.300} & 0.192 & 0.138 & -0.163 & -0.176 \\
        \hline
        \textit{Yeah, and to save you from any embarrassment umm, I think maybe I should talk first.} & 0.245 & 0.100 & 0.311 & \textbf{0.381} & -0.026 & 0.126 \\
        \hline
        \textit{Hey. Call me when you get there. Okay?} & 0.143 & 0.206 & 0.158 & \textbf{0.365} & -0.049 & 0.314 \\
        \hline
        \textit{What?! I didn't touch a guitar!} & 0.407 & 0.293 & \textbf{0.646} & 0.529 & 0.284 & 0.320 \\
        \hline
    \end{tabular}
    \caption{Pearson correlations ($r$) between human category annotations and category encodings produced by traditional lexicon-based methods, \ours, and \oursp. We use \oursboth with both pre-trained and fine-tuned MPNet as $M_\theta$.}
    \label{tab:topic_corrs}
\end{table*}

\end{document}